# CRENER: A Character Relation Enhanced Chinese NER Model

Yaqiong Qiao[a,*], Shixuan Peng[b]

[a] Nankai University,Tianjin, China 450046

[b] School of Information Engineering, North China University of Water Resources and Electric Power, China 450046

# Abstract

Chinese Named Entity Recognition (NER) is an important task in information extraction, which has a significant impact on downstream applications. Due to the lack of natural separators in Chinese, previous NER methods mostly relied on external dictionaries to enrich the semantic and boundary information of Chinese words. However, such methods may introduce noise that affects the accuracy of named entity recognition. To this end, we propose a character relation enhanced Chinese NER model (CRENER). This model defines four types of tags that reflect the relationships between characters, and proposes a fine-grained modeling of the relationships between characters based on three types of relationships: adjacency relations between characters, relations between characters and tags, and relations between tags, to more accurately identify entity boundaries and improve Chinese NER accuracy. Specifically, we transform the Chinese NER task into a character-character relationship classification task, ensuring the accuracy of entity boundary recognition through joint modeling of relation tags. To enhance the model's ability to understand contextual information, WRENER further constructed an adapted transformer encoder that combines unscaled direction-aware and distance-aware masked self-attention mechanisms. Moreover, a relationship representation enhancement module was constructed to model predefined relationship tags, effectively mining the relationship representations between characters and tags. Experiments conducted on four well-known Chinese NER benchmark datasets have shown that the proposed model outperforms state-of-the-art baselines. The ablation experiment also demonstrated the effectiveness of the proposed model.

---

[*] Corresponding author. Email address: kitesmile@126.com

# 1. Introduction

Named entity recognition (NER) is a fundamental task of natural language processing (NLP) [1], and plays a crucial role in various downstream tasks, such as information retrieval [2], entity linking [3], and relationship extraction [4][5].

However, Chinese NER faces significant challenges due to the absence of word separators and the complex structures of named entities, which include nested and discontinuous entities [6][7]. The ambiguous division of character boundaries can lead to errors in entity segmentation, subsequently causing incorrect entity classification [8], thereby significantly elevating the difficulty of the Chinese NER task.

Most of the studies [9][10][11] treat NER as a sequential labeling problem, assigning labels to tokens based on their entity types. The state-of-the-art methods can be grouped into the following categories: sequence-to-sequence (Seq2Seq)-based method [12], Hypergraph-based method [13], span-based method [14], Large Language Model (LLM)-based method [15] and grid-tagging-based method [16].

Among these methods, the first four methods all have certain limitations. Seq2Seq-based methods suffer from inefficient inference and error propagation, hindered by sequential dependency in capturing and labeling entities. Hypergraph-based methods struggle with structural errors and error propagation during prediction due to the gradual graph generation process. Span-based methods are constrained by maximum span length and high computational complexity, limiting their scalability for longer sequences. LLM-based methods demand extensive labeled data, high computational resources, and may mislabel non-entities as named entities in lower-level language comprehension tasks. On the contrary, grid-tagging-based methods performed relatively well compared to the first four methods, which enhance entity extraction accuracy by predicting relation matrices for character pairs within sentence structures.

In this paper, we carried out NER research based on grid-tagging methods. Although grid-tagging methods performs better compared to other methods, our research has also identified some defects of this approach.

For example, Li et al. [16] designed a multi-granularity 2D convolution to improve the word pair representations, and used a co-predictor to reason the word-word relations. Their framework and model are easy to migrate, but they only use two tags, resulting in a sparse distribution of tags in the grid, it cannot effectively handle the character-character relationships for different types of entities in Chinese NER. Liu et al. [17] designed a Tag Representation Embedding Module with four tags to model the relationships among words and tags. Their model can better identify discontinuous entities, but they only conducted research on discontinuous entity recognition on three English datasets. Additionally, the aforementioned grid-tagging based methods all adopt Bi-directional Long Short-Term Memory (BiLSTM) to generate word representations, resulting in insufficient efficiency due to its sequential input architecture [18].

Therefore, to enhance the efficiency and accuracy of Chinese NER, we propose a character Relation Enhanced NER Model (CRENER). Specifically, we employ four- character embedding strategies to generate the semantic representation of sentences, and construct an adapted Transformer encoder to model the character-level representation, which combines the direction-aware and distance-aware masked self-attention to extract global context information. Furthermore, we adopt the Conditional Layer Normalization (CLN) to generate the representation of character–character grids, and construct a convolution module to capture the interaction information between characters with different distances. Moreover, we employ four tags to model fine-grained character-character relationships, and construct a relation enhancement module to embed the tag representation between characters into the model. Finally, we jointly predict the fused character and tag representations through a co-predictor module, and named entities can be decoded from all possible entity mentions.

The main contributions of this paper are as below:

1. We innovatively construct an adapted Transformer encoder that combines unscaled direction-aware and distance-aware masked self-attention mechanisms for global context encoding, enhancing the model's capacity for contextual understanding.
2. We develop a novel relation enhancement module to model four predefined relation tags in the 2D grid, capturing interactions between characters with different distances, enhancing entity prediction accuracy, and significantly improving the representation of such relations within the model.
3. we propose a character Relation Enhanced NER Model (CRENER). Experiments conducted on four well-known Chinese NER benchmarking datasets verified the superiority of it.

The rest of this paper is organized as follows. Section 2 presents the related works about Chinese NER; Section 3 details the proposed model; Section 4 introduces the datasets used in this paper, and conducts experiments to analyze our model; Finally, we give conclusions in Section 5.

## 2. RELATED WORK

Most of the early Chinese NER methods are rule-based or statistics-based. The rule-based methods require domain-specific experts to manually construct the rule templates[19], and the process of creating rules is significantly expensive due to the variety of features involved. The statistics-based methods mainly use Conditional Random Fields (CRF) or the hidden Markov model (HMM) for training the NER model [20]. They are not able to retain inherent semantic information during the process of NER, resulting in low entity recognition accuracy [21].

With the development of deep neural networks and pre-trained language models, some new methods have emerged. These methods can be categorized into the following five types: Seq2Seq-based method [12] [22], Hypergraph-based method [13][23] [24], span-based method [14][25][26], LLM-based method [15] and grid-tagging-based method [7][16][17].

The Seq2Seq-based method generates the entity index sequences based on the Encoder-Decoder framework. Many studies devised various translation schemas to unify the NER task with text generation [22]. For instance, Yan et al.[12] formulate the NER subtasks as an entity span sequence generation task, leveraging pre-trained Seq2Seq models and three entity representations to solve all subtasks without the special design of the tagging schema or ways to enumerate spans. This kind of method faces issues such as disordered entity sequences and incorrect decoding biases [27].

Span-based methods identify Chinese-named entities by recognizing and classifying continuous text spans using models that can predict the starting and ending positions of these spans within the text [28]. For instance, Fu et al. [25] employed a span-based constituency parser to handle nested NER and eliminated the error propagation problem using globally exact inference based on the masked inside algorithm. This kind of method offers greater flexibility and accuracy in recognizing overlapping and nested entities[14], but it confronts challenges related to decoding efficiency and exposure bias[27].

The hypergraph-based method represents all entity spans using a hypergraph, capturing complex relationships among entities, and learning to combine graph nodes with individual classifiers [17]. For instance, Wang and Lu [13] utilized a novel segmental hypergraph representation to model overlapping entity mentions in text, capturing features and interactions previously unattainable while maintaining time complexity. This kind of method faces challenges such as spurious structures, structural ambiguity, and susceptibility to exposure bias [29].

The LLM-based method leverages the contextual understanding and prediction capabilities of large language models to identify and classify named entities within text. For instance, Lou et al. [15] introduce an in-context learning NER method using PLMs modeled as meta-functions, pre-trained with instructions and demonstrations, to recognize novel entity types with limited examples, surpassing traditional fine-tuning. This kind of method requires significant computational resources and struggles with domain-specific or specialized entities not well-represented in its training data [30].

Grid-tagging-based methods entail constructing a grid and extracting entities by predicting the relation matrix between words [31]. For instance, Li et al. [16] used BiLSTM to generate the final word representation and multi-granularity 2D convolutions to extract the relationship between characters for prediction. Liu et al. [17] extended the tagging system with two additional tags to model word relationships and reduced error propagation in tagging discontinuous entities. This kind of method utilizes a 2D representation to capture word relationships and entity spans through a simpler end-to-end process, thereby minimizing error propagation and avoiding the drawbacks associated with other approaches [32].

Inspired by [16] [17], we focus on mining the relationship among characters and tags to iteratively optimize the convolutional module inputs, leveraging a more fine-grained tagging system to strengthen the co-predictor module's prediction of the relationships between characters and tags.

## 3. Preliminary

For the input sentence consists of N characters and a predefined entity type set consists of M entity types, the goal of this paper is to extract a set of entities from $X$ with their corresponding entity type. In this paper, we use $X = \{x_1, x_2, ..., x_N\}$ denotes the input sentence, $Y = \{y_1, y_2, ..., y_M\}$ denotes the predefined entity type set, $E = \{e_1^{y_1}, e_2^{y_2}, ..., e_p^{y_p}\}$ denotes the set of entities extract from X, and use a two-dimension matrix L to represent the output of the proposed method. The Chinese NER model can be formulated as below,

$$L = \text{CRENER}(X, Y)$$

where $x_i$ is a token denoting to a character, $y_i$ is a kind of predefined entity type, $e_i^{y_i}$ is a specific entity. $L_{ij}$ in $L$ means the entity starts from $x_i$ to $x_j$.

We transform the Chinese NER task into a character-character relation tag prediction task. The adjacency relationship between characters is transformed into a 2D grid representation, and entity recognition is extracted by predicting the tags between characters. we provide a detailed introduction to the tags used in our model as follows:

- Next-Neighbor-Character (NNC) indicates that the character pair $(x_i, x_j)$ belongs to the same entity, the character $x_i$ in a specific row in the upper triangle of the grid has a continuous $x_j$ in a specific column;
- Previous-Neighbor-Character (PNC) indicates that the character pair $(x_i, x_j)$ belongs to the same entity, the character $x_j$ in a specific row in the lower triangle of the grid has a continuous $x_i$ in a specific column;
- Tail-Head-Character (THC) indicates that the character $x_i$ in a specific row is the Tail of the entity, and the character $x_j$ in the grid column is the Head of the entity.
- Head-Tail-Character (HTC) indicates that the character $x_i$ in the grid row is the Head of the entity, and the character $x_j$ in the grid column is the Tail of the entity.
- None: It means that the character pair $(x_i, x_j)$ has no relation.

With the above four predefined tags, the sentence $X$ can be represented as a grid and decoded to get all entities. Figure 1 shows a concrete example.

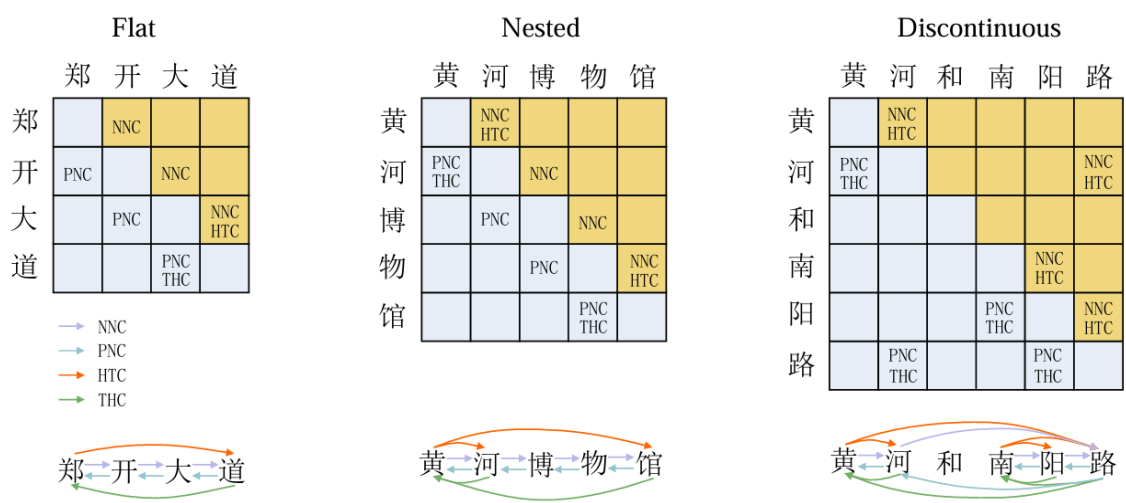

Figure 1 Three types of entities are represented by five predefined tags, include NONE, NNC, PNC, HTC and THC. The relation grids demonstrate the character–character relation modeling method, which can be transformed into the directed graphs below.

Table 1 List of notations

| Notations | Description |
|-----------|-------------|
| $H$ | The representation of the input sentence |
| $H_s, H_o$ | The directional character representation of subject and object |
| $V_{ij}$ | The grid matrix element that is defined by character pair $(x_i, x_j)$ |
| $V$ | The character pair representation grid by CLN |
| $E^d, E^r$ | The distance matrix and region matrix |
| $Q$ | The interaction matrix between characters with different distances |
| $TF_t(i,j)$ | Tag-aware feature of character pairs $(x_i, x_j)$ with dilation rates $t$ |
| $TF^{(t)}$ | The concatenated tag-aware feature at the $t$-th iteration |
| $H_{s(ll)}^{(t)}, H_{o(ll)}^{(t)}$ | The tag-aware character representation of subject and object |
| $H_{s(Wl)}^{(t)}, H_{o(Wl)}^{(t)}$ | The relation representation between $H_s, H_o$ and $H_{s(ll)}^{(t)}, H_{o(ll)}^{(t)}$ |
| $y_{ij}$ | The output of co-predictor for character pairs $(x_i, x_j)$ |

# 4. Proposed Model

We formulate the named entity recognition task as a grid tagging problem. In this section, all possible entities are identified by predicting the character pair relation grid corresponding to the input sentence using four predefined tags. The proposed model includes four main components: encoder module, convolution module, relation enhancement module, co-prediction module, and Decoding. The model architecture is shown in Figure 2.

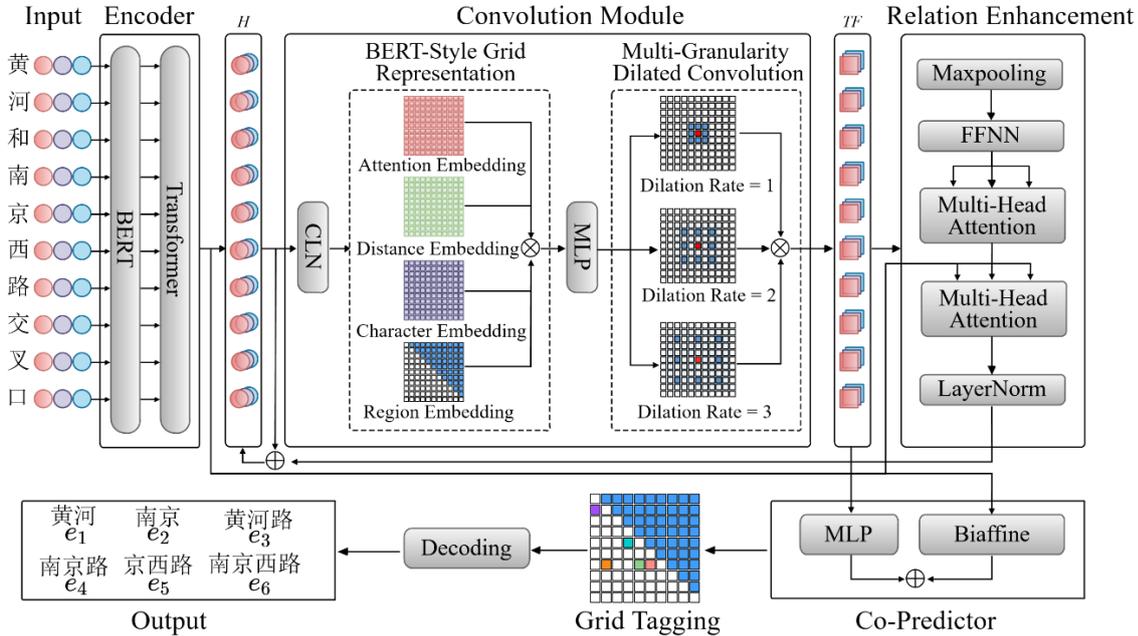

Figure 2 Overall structure of the proposed model. $\otimes$ represents concatenation operations and $\oplus$ represents element-wise addition. $H$ represents character representation and $TF$ represents tag-aware grid features.

## 4.1 Encoder

### 4.1.1 Semantic feature extraction

We extract semantic features using four character embedding strategies [33]: BERT, distance, region, and attention embeddings. Specifically, we utilize the pre-trained language model BERT[34] to obtain the semantic features of characters. The distance embedding is used to capture the positional features of characters within a sentence. The region embedding is used to distinguish the upper and lower triangle regions of a matrix. The attention embedding is derived from the raw input through an attention layer, which assigns weights to input features based on their relevance to the output.

We represent these word embeddings in order as follows: $H^B = \{h_1^B, h_2^B, ..., h_N^B\}$, $H^D = \{h_1^D, h_2^D, ..., h_N^D\}$, $H^R = \{h_1^R, h_2^R, ..., h_N^R\}$, $H^A = \{h_1^A, h_2^A, ..., h_N^A\}$, respectively.

### 4.1.2 Context modeling enhancement

Since the Transformer model uses a fully connected self-attention mechanism structure to extract global context information, it is far superior to the recurrent neural network in parallel computing[35]. Nevertheless, the scaled and smooth attention distribution of the vanilla Transformer [36] may contain some noisy information, and the information from different representations at different positions is easy to ignore. Therefore, to further enhance context modeling, we employ an adapted Transformer encoder that combines unscaled direction-aware and distance-aware masked self-attention mechanisms to model character-level features. The structure of the adapted Transformer model is shown in Figure 3.

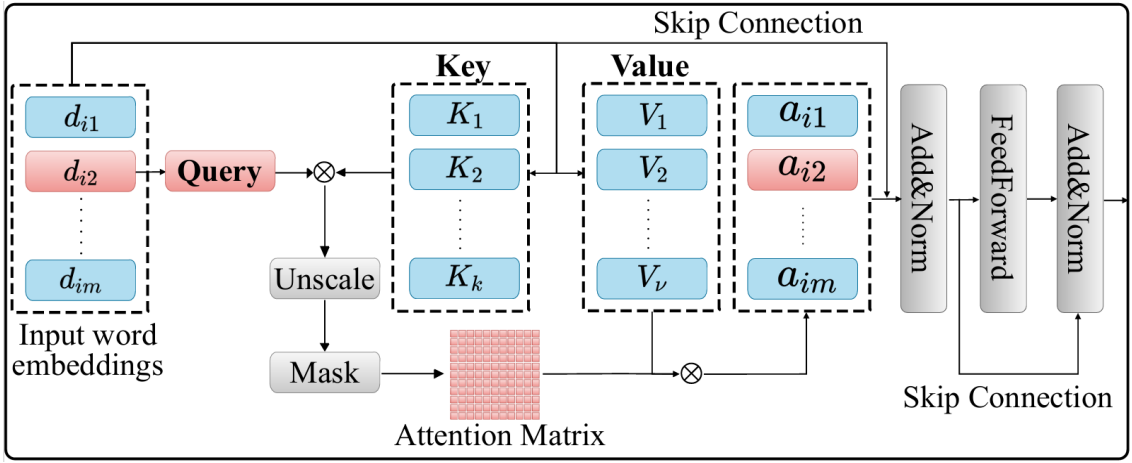

Figure 3 The structure of the adapted Transformer

The structure can be described as mapping a query from a set $\{d_{i1}, d_{i2}, ..., d_{im}\}$ to an output $\{a_{i1}, a_{i2}, ..., a_{im}\}$ using a set of keys $\{K_1, K_2, ..., K_k\}$ and a corresponding set of values $\{V_1, V_2, ..., V_v\}$, where the query, keys, values, and output are all vectors. The output is computed as a weighted sum of the values, where the weights are determined by a compatibility function calculated using a fully connected feed-forward network based on the query and its corresponding key. Additionally, we employ the residual connection to improve the flow of gradients during training.

The details of the Direction-aware and Distance-aware Masked Self-Attention are described as follows: given an input sequence representation $X = \{x_1, x_2, ..., x_N\}$, we can first transform it into queries $Q = H_i W_q$, keys $K = H_i W_k$, and values $V = H_i W_v$, where $W_q, W_k, W_v$ are learnable parameters. We use sine and cosine functions to calculate the relative positional embedding.

$$R_{i,j} = [...\sin(\frac{d_{i,j}}{10000^{2k/d_{\text{model}}}}) \cos(\frac{d_{i,j}}{10000^{2k/d_{\text{model}}}})...]^T \quad (1)$$

where $d_{i,j}$ represent the relative distance of the target token $i$ and the context token $j$, $d_{\text{model}}$ is the dimension of position encoding and k is the index of position encoding. It has no learnable parameters. Then the weights of self-attention are calculated as follows:

$$\begin{aligned} A_{i,j}^{rel} &= W_q^T H_i^T H_j W_k + W_q^T H_i^T R_{i,j} W_{kR} + u^T H_j W_k + v^T R_{i,j} \\ &= Q_i^T K_j + Q_i^T R_{i,j} + u^T K_j + v^T R_{i,j} \end{aligned} \quad (2)$$

$$\text{Attn}(Q, K, V) = \text{softmax}(A_{i,j}^{rel})V \quad (3)$$

where $u, v$ are learnable parameters. The attention score computation $A_{i,j}^{rel} = f(Q_i, K_j)$ obtain the attention score between query vector $Q_i$ of token $i$ and key vector $K_j$ of token $j$. The performance of $A_{i,j}^{rel}$ in Eq.(3) without the scaling factor $\sqrt{d_k}$ surpasses the vanilla Transformer, presumably because the absence of the scaling factor sharpens the attention, benefiting the NER task where only a few words are named entities.

After encoded by transformer, the input sentence $X = \{x_1, x_2, ..., x_N\}$ will be final represented as: $H = \{h_1, h_2, ..., h_N\} \in R^{N \times d_h}$, $N$ denotes the length of $X$, $d_h$ denotes the dimension of the character vector, and $h_i = \text{Concat}(h_i^B \oplus h_i^D \oplus h_i^R \oplus h_i^A) \in R^{d_h}$ is the representation of the $i$-th character.

## 4.2 Convolution Module

The convolution module comprises three main components: a conditional layer normalization (CLN) [37] that generates the representations of character-pair grids, a bert-style grid representation that exhibits the relationships between character pairs, and a multi-granularity dilated convolution that captures the interactions among characters with different distances.

As the relationships between character pairs in this paper are directional, with each character pair potentially serving as the head, tail, or middle segment of an entity in a relationship, we aim to model these relationships between character pairs, as shown in the figure 1. The relationship between the subject and object character representations in our entity is expressed as follows:

$$h_i^s = W_h^s h_i + b_h^s,$$
$$h_i^o = W_h^o h_i + b_h^o,$$
(4)

where $h_i^s, h_i^o \in R^{d_h}$ denotes the subject and object representations of the $i$-th character, $W_h^* \in R^{d_h \times d_h}$ and $b_h^* \in R^{d_h}$ are trainable weights and biases respectively.

### 4.2.1 Conditional layer normalization

We first generate the representation of characters on the grid by CLN, which can be viewed as a three-dimensional matrix $V \in R^{N \times N \times d_h}$. Specifically, For each pair of characters $(x_i, x_j)$, the grid is defined by row elements $x_i$ and column elements $x_j$. Each matrix element $V_{ij}$ in $V$ represents the interaction between character representations $h_i^s$ of $x_i$ and $h_j^o$ of $x_j$, which $x_i$ can be considered as a condition for $x_j$. The CLN is formalized as follows:

$$V_{ij} = CLN(h_i^s, h_j^o) = \gamma_{ij} \odot (\frac{h_j^o - \mu}{\sigma}) + \lambda_{ij}$$
(5)

where $h_i^s$ represents the condition to generate the gain parameters $\gamma_{ij} = W_\alpha h_i^s + b_\alpha$ and bias $\lambda_{ij} = W_\beta h_j^o + b_\beta$ in the layer normalization. $W_\alpha, W_\beta \in R^{d_h \times d_h}$ and $b_\alpha, b_\beta \in R^{d_h}$ are trainable weights and biases. $\mu$ and $\sigma$ are mean and standard deviation taken across the elements of $h_j^o$:

$$\mu = \frac{1}{d_h} \sum_{k=1}^{d_h} h_{jk}^o, \sigma = \sqrt{\frac{1}{d_h} \sum_{k=1}^{d_h} (h_{jk}^o - \mu)^2}$$
(6)

where $h_{jk}^o$ is the element of the $k$-th dimension of $h_j^o$.

### 4.2.2 Bert-style grid representation

To enrich the representation of the character-pair grid, we establish a distance matrix $E^d \in R^{N \times N \times d_{E_d}}$, a region matrix $E^r \in R^{N \times N \times d_{E_r}}$ and a attention matrix $E^a \in R^{N \times N \times d_{E_a}}$, where the dimensions of $E^d, E^r$ and $E^a$ are $d_{E_d}, d_{E_r}$ and $d_{E_a}$ respectively. $E^d$ denotes the relative distance between character pairs, $E^r$ combines directional information and distinguishes the upper and lower triangular areas, $E^a$ denotes the weights representing the relevance of input features to the output. Then it is mixed with character pair information $V \in R^{N \times N \times d_h}$, and the location region perception representation is obtained by multi-layer perceptron (MLP) dimension reduction:

$$C = MLP_1(V \otimes E^d \otimes E^r \otimes E^a) \quad (7)$$

### 4.2.3 Multi-granularity dilated convolution

We capture the interaction information between characters with different distances by controlling the dilation rate of the multi-granularity dilated convolution [16]. In this paper, we adopt multiple 2-dimensional(2D) dilated convolutions (DConVs) with dilation rates $\iota \in [1,2,3]$, and the grid representation is $Q = (Q^1 \otimes Q^2 \otimes Q^3) \in R^{N \times N \times 3d_c}$, the formula is:

$$Q^\iota = GELU(DConv_\iota(C)) \quad (8)$$

where $Q^\iota \in R^{N \times N \times d_c}$ is the output when the dilation rate is $\iota$.

## 4.3 Relation Enhancement Module

To model the interaction information among characters and tags, we construct the relation enhancement module to embed the relation representation among characters and tags into the model. To capture the tag-aware grid features, we align the number of tags by transforming the dimensionality of the grid representation $Q^\iota$ of character pairs $(x_i, x_j)$. The tag-aware feature is formalized as:

$$TF_\iota(i,j) = W_\iota Q_{ij}^{\ \iota} + b_\iota \quad (9)$$

where $TF_\iota(i,j)$ represents the tag-aware grid features of elements $(i,j)$ in character pairs $(x_i, x_j)$, $W_\iota \in R^{d_r \times d_c}$ and $b_\iota \in R^{d_r}$ are trainable weights and biases respectively.

Since the information from different representations at different positions is easy to ignore, using only a single attention head will inhibit information from different representation subspaces at different positions. As we perform entity prediction through joint extraction of relations among tags, and subsequently concatenate four kinds of tags we used in our paper together as below:

$$TF^{(r)} = Concat(TF_{NNC}^{(r)} \otimes TF_{PNC}^{(r)} \otimes TF_{HTC}^{(r)} \otimes TF_{THC}^{(r)}) \quad (10)$$

where r represents the relation enhancement module runs several rounds to optimize $TF \in R^{N \times N \times 4d_r}$.

We feed the input $TF^{(r)}$ from different dimensions into two separate Max pooling layers ($Maxpool_1, Maxpool_2 \in R^{N \times 4d_r}$) and the Feed-Forward Network (FFN) layer to recover the tag-aware feature to the subject and object character features $H_s^{(r)}$ and $H_o^{(r)}$ at the $r$-th iteration:

$$\begin{aligned} H_s^{(r)} &= S-FFN(Maxpool_1(TF^{(r)})W_s + b_s), \\ H_o^{(r)} &= O-FFN(Maxpool_2(TF^{(r)})W_o + b_o). \end{aligned} \quad (11)$$

where $W_s, W_o \in R^{4d_r \times d_h}$ and $b_s, b_o \in R^{d_h}$ are trainable weights and biases respectively. $Maxpool_1$ and $Maxpool_2$ merge the tag representations $TF^{(r)}$ with the row elements $x_i$ and column elements $x_j$ of the table respectively, in order to recover the representations of the subject characters $H_s^{(r)}$ and object characters $H_o^{(r)}$.

Since the information from different representations at different positions is easy to ignore, using only a single attention head will inhibit information from different representation subspaces at different positions. We utilize a multi-head self-attention mechanism [39], enabling the model to simultaneously focus on different representation subspaces at different positions. We fed the recovered character features $H_s^{(r)}$ and $H_o^{(r)}$ concurrently as Query, Key, and Value into the multi-head self-attention mechanism, so as to extract relationships between these tag-aware character representations.

$$H_{s(tt)}^{(r)} = SelfAttention(H_s^{(r)}, H_s^{(r)}, H_s^{(r)}),$$
$$H_{o(tt)}^{(r)} = SelfAttention(H_o^{(r)}, H_o^{(r)}, H_o^{(r)}). \quad (12)$$

Subsequently, we take the output of the previous round of attention $H_{s(tt)}^{(r)}$ and $H_{o(tt)}^{(r)}$ as the Query, and the character representation $H_s$ and $H_o$ as the Key and Value. Meanwhile, we send it to another multi-head cross-attention mechanism to learn the relationships between character representations and the tag-aware character representations:

$$H_{s(ct)}^{(r)} = CrossAttention(H_{s(tt)}^{(r)}, H_s, H_s),$$
$$H_{o(ct)}^{(r)} = CrossAttention(H_{o(tt)}^{(r)}, H_o, H_o). \quad (13)$$

Then, we apply a linear transformation to the output of the cross-attention mechanism and used a GELU activation function to learn more complex feature representations:

$$H_{s(ct)}^{(r)} = GELU(W_s H_{s(ct)}^{(r)} + b_s),$$
$$H_{o(ct)}^{(r)} = GELU(W_o H_{o(ct)}^{(r)} + b_o). \quad (14)$$

where $W_s, W_o \in R^{4d_r \times d_h}$ and $b_s, b_o \in R^{d_h}$ are trainable weights and biases respectively.

During the iterative optimization process where the relation enhancement module refeeds its output back into the convolution module, the tag-aware character representations encounter the potential issue of gradient vanishing. To address this, we adopt residual connections [40] to ensure smoother gradient flow, thereby enhancing the module's performance.

$$H_s^{(r+1)} = LayerNorm(H_s^{(r)} + H_{s(ct)}^{(r)}),$$
$$H_o^{(r+1)} = LayerNorm(H_o^{(r)} + H_{o(ct)}^{(r)}). \quad (15)$$

## 4.4 Co-Predictor module

After obtaining tag-aware grid features for each character pair through the relation enhancement module, the MLP receives these features to predict the relationships between the character pairs. In addition, previous studies have demonstrated that combining the MLP predictor with the biaffine predictor can enhance the classification of relation [4] Therefore, we simultaneously use both predictors to calculate two separate relation distributions for character pairs $(x_i, x_j)$ and combine them to generate the final prediction.

The computation in the biaffine predictor can be described as follows: for each character pair $(x_i, x_j)$, we utilize two MLPs to compute the character representations $s_i$ and $o_j$ for $x_i$ and $x_j$, and the biaffine classifier to calculate the relation score between them. This process can be described as follows:

$$s_i = MLP(h_i), o_j = MLP(h_j),$$
$$y'_{ij} = s_i^T U o_j + W[s_i; o_j] + b \qquad (16)$$

where U, W, and b are parameters to learn, $s_i$ and $o_j$ represent the subject and object representations of the *i*-th and *j*-th character respectively, and $y'_{ij} \in R^{|\Re|}$ is the prediction scores of the predefined relations of character pairs $(x_i, x_j)$.

Based on the output $TF^{(N)}$ of the relation enhancement module, we use an MLP to calculate the relation score for each character pair $(x_i, x_j)$:

$$y''_{ij} = MLP(TF^{(N)}(i, j)) \qquad (17)$$

where $y''_{ij} \in R^{|\Re|}$ is the prediction scores of the predefined relations of character pairs $(x_i, x_j)$.

Finally, the prediction scores of the biaffine predictor $y'_{ij}$ and the MLP predictor $y''_{ij}$ are combined to calculate the final probability distribution of the character pair $(x_i, x_j)$, and the relationship of the character pair $(x_i, x_j)$ is determined by the maximum value in $y_{ij}$:

$$y_{ij} = Soft\max(y'_{ij} + y''_{ij}) \qquad (18)$$

## 4.5 Training

The above describes the forward calculation of the proposed architecture. Since there may be more than one relationship between each character pair, for each sentence $X = \{x_1, x_2, ..., x_N\}$, the training goal is to predict the correct tag. So we define a threshold to filter the target tags, simultaneously ensure the score of each target tag is not less than the score of each non-target tag. We adopt a cross-entropy loss function for multi-tag classification [17], formalized as follows:

$$\mathcal{L} = \log(1 + \sum_{n \in \Omega_{neg}} e^{s^n_{(i,j)}} \sum_{m \in \Omega_{pos}} e^{-s^m_{(i,j)}} + \sum_{n \in \Omega_{neg}} e^{s^n_{(i,j)} - s_0} + \sum_{m \in \Omega_{pos}} e^{s_0 - s^m_{(i,j)}})$$
$$= \log(e^{-s_0} + \sum_{m \in \Omega_{pos}} e^{-s^m_{(i,j)}}) + \log(e^{s_0} + \sum_{n \in \Omega_{neg}} e^{s^n_{(i,j)}}) \tag{19}$$

where $\Omega_{pos}$ and $\Omega_{neg}$ are the target and non-target tag sets respectively. $s^m_{(i,j)}$ and $s^n_{(i,j)}$ are the target and non-target tag scores respectively. $s_0$ represents the threshold.

## 4.6 Decoding

For the decoding process, our final goal is to find all character sequences of entities and corresponding entity types. By utilizing four predefined tags, we can model fine-grained character-character relations and compensate for some error propagation in model predictions. Furthermore, for the input sentence $X = \{x_1, x_2, ..., x_N\}$, the model outputs the characters and their relation tags. The pseudo code for the CRENER model is shown in Algorithm1.

For example, we first iteratively find all THC and HTC relationships in the lower triangle of the grid. As entities are not independent of each other, THC and HTC relations may correspond to one or more entities. If the entity contains a single character, only THC and HTC relations are used to decode it. For multi-character entities, the entire grid corresponding to the sentence is converted into a directed graph, and when both co-predicted NNC and PNC relationships are present, we believe that character pairs belong to the same entity, as shown in the figure 1. In this graph, nodes represent characters and edges represent four kind of relations. The depth first search algorithm is used to find all paths from the head character to the tail character, which is the character sequence of the entity.

---

**Algorithm1:** Pseudo code of CRENER model

**Input:** A matrix of relations $R$ for a sentence $X$, where $R_{ij}$ represents the relation between character $i$ and character $j$, with $i,j \in [1,N]$.

**Output:** A list of entities with their character index sequence set $E$ and label set $L$.
    1: Initialize entity sets and tag sets $E = []$, $T = []$.
    2: Obtain the Chinese representations of character embeddings using four-character embedding strategies.
    3: Fuse the Chinese representation embeddings and generate representations of the character pair grid $(x_i, x_j)$ using equations (4) to (8).
    4: Enhance the relation representation among characters and tags using equations (9) to (15).
    5: for $R_{ij} \in R$ and $i \geq j$ do
    6:    if $R_{ij} \in (HTC)$ relation or $R_{ij} \in (THC)$ relation then
    7:       Create a sequence $S \leftarrow [j]$
    8:       if $i=j$ then

```
9:            Add S to E
10:           Add corresponding tag t to T
11:      else
12:           for k ∈ (j, N] do
13:               Search(S, R_{jk}, R_{kj}, k, i, R_{ij})
14: return E, T
```
```
15: Function Search(S, r_1, r_2, m, n, t):
16: if r_1 ∈ (NNC) relation and r_2 ∈ (PNC) relation then
17:      Add m to S
18:      if m=n then
19:           Add S to E
20:           Add corresponding tag t to Tss
21:      else
22:           for k ∈ (m, N] do
23:               Search(S, R_{mk}, R_{km}, k, n, t)
```

# 5. Experiments

We conduct experiments on four mainstream Chinese NER benchmark datasets and conduct ablation experiments to detail our experimental results and experimental details. Standard precision, recall, and F1 score are used as evaluation metrics. The experimental results show that the proposed model has better performance.

## 5.1 Datasets

We also use four well-known Chinese NER datasets, including (1) Weibo [41] (2)Resume [42] (3) Ontonotes 4.0 [43] (4) MSRA [44]. These four datasets come from different fields, and their writing forms are also very different. Among them, the corpus of Weibo and Resume are from social media and Sina Finance, and there is no benchmark word segmentation on these two datasets. While MSRA and Ontonotes 4.0 are from news, whose benchmark word segmentation is available for training data. For OntoNotes, benchmark word segmentation is also available for development and test data. The statistics of the four datasets are shown in Table 1.

Table 1 Statistics of the benchmarking datasets.

| Datasets | Types | Train | Dev | Test | Entity Types |
|---|---|---|---|---|---|
| Weibo | Sentences | 1.35k | 0.27k | 0.27k | 4 |
|  | Entities | 1.89k | 0.39k | 0.42k |  |
| Resume | Sentences | 3.8k | 0.46k | 0.48k | 8 |
|  | Entities | 1.34k | 0.16k | 0.15k |  |
| OntoNotes | Sentences | 15.7k | 4.3k | 4.3k | 4 |
|  | Entities | 13.4k | 6.95k | 7.7k |  |
| MSRA | Sentences | 46.4k | - | 4.4k | 3 |
|  | Entities | 74.8k | - | 6.2k |  |

## 5.2 Baselines

We compare our model with multiple NER models on different datasets, depending on whether the source code is publicly available or not. All the baselines are derived from published papers.

- CAN-NER [45] combines CNN with a local attention mechanism and uses small character embeddings without relying on any external resources, making CAN-NER more practical in practical system scenarios.
- softLexicon [46] introduces lexical information with only minor adjustments to the representation layer of characters.
- MSFM [47] combine multi-dimensional features to improve the recognition ability of Chinese sentence entities.
- MECT [48] integrates the structural information of Chinese characters with multivariate data embedding cross Transformer, which can better capture the semantic information of Chinese characters.
- NER-MC [19] combines word boundary information with semantic information to improve the performance of entity recognition.
- W²NER [16] employs a grid-token-based approach to assign a token to each pair of words from which entities can be decoded.
- Token-Relation [20] proposes a hidden self-attention mechanism to incorporate the semantics of latent words into their local context information.
- VisPhone [49] fuses visual and speech features of input characters with text embeddings and adopts a selective fusion module to obtain the final features.
- DAE-NER [50] designs attention enhancement modules on characters and sentences to obtain the semantic representation information of characters with different granularities in the text.
- MFT [51] improves the basic structure of the Transformer model, further enhances the semantic information by adding the word root information of Chinese characters, and achieves good performance on resume and weibo datasets.

## 5.3 Results and Analyses

Weibo NER dataset: Table 2 shows the results obtained on the Weibo dataset. The F1 of CAN-NER(Zhu et al.,2019), MSFM were 59.31% and 55.94%, respectively. The F1 score of DAE-NER with the attention enhancement module is 57.45%. The F1 of MECT and MFT of Chinese character glyph feature is 63.30% and 64.38% respectively. The F1 score of VisPhone with speech features added to glyphs is 70.79%. We can find that our model improves F1-score and recall by 4.31% and 5.56% respectively compared with W²NER, and improves precision by 1.21% compared with Token-Relation. The above experimental results prove that our model is the best one compared with other models.

Table 2 Results obtained on Weibo.

| Models | Chinese Weibo NER | | |
| --- | --- | --- | --- |
|  | Precision | Recall | F1 |
| CAN-NER(Zhu et al.,2019) | 55.38 | 62.98 | 59.31 |
| SoftLexicon (Ma et al. 2020) | 70.94 | 67.02 | 70.50 |
| MECT (Wu et al., 2021) | 61.91 | 62.51 | 63.30 |
| MSFM (Liu et al., 2022) | 60.75 | 51.83 | 55.94 |
| MFT (Han et al., 2022) | 63.72 | 65.03 | 64.38 |

| Models | | | |
|---|---|---|---|
| W²NER(Fei et al,2022) | 70.84 | 73.87 | 72.32 |
| Token-Relation(huang et al,2022) | 72.82 | 66.02 | 69.62 |
| NER-MC(Yan et al,2023) | 62.20 | 64.05 | 63.06 |
| VisPhone (Zhang, B., et al.2023） | 65.65 | 71.29 | 70.79 |
| DAE-NER（Sun et al., 2024) | 69.68 | 48.89 | 57.45 |
| ours | **74.03** | **79.43** | **76.63** |

Resume NER dataset: Table 3 shows the results obtained on the Resume dataset. The F1 scores of MECT (Wu et al., 2021) and NER-MC(Yan et al,2023) fused with semantic information are 95.89% and 95.16%. SoftLexicon (Ma et al., 2019) and MFT (Han et al., 2022) introduced incorporating lexical information into characters to enhance semantics, with F1 scores of 96.11% and 95.78%, respectively. CAN-NER (Zhu et al., 2019) and Token-Relation(huang et al,2022) obtained f1 scores of 95.74% and 96.36% by using an improved attention mechanism. The respective F1 of MSFM are 95.43%. The F1 value of VisPhone with speech features added on the basis of glyph is 96.26%, and the F1 value of DAE-NER with attention enhancement module is 96.04%. Our model achieves the highest F1 value, precision, and recall, which are 96.86%, 97.16%, and 96.56%, respectively. Compared with W²NER, the F1 value and precision are increased by 0.31% and 0.20%, respectively, and the optimal score is achieved.

Table 3 Results obtained on Resume.

| Models | Chinese Resume NER | | |
|---|---|---|---|
| | Precision | Recall | F1 |
| CAN-NER(Zhu et al.,2019) | 95.71 | 95.77 | 95.74 |
| SoftLexicon (Ma et al. 2020) | 96.08 | 96.13 | 96.11 |
| MECT (Wu et al., 2021) | 96.40 | 95.39 | 95.89 |
| MSFM (Liu et al., 2022) | 96.08 | 94.79 | 95.43 |
| MFT (Han et al., 2022) | 96.05 | 95.52 | 95.78 |
| W²NER(Fei et al,2022) | 96.96 | 96.35 | 96.65 |
| Token-Relation(huang et al,2022) | 96.01 | 96.50 | 96.36 |
| NER-MC(Yan et al,2023) | 94.60 | 95.73 | 95.16 |
| VisPhone (Zhang, B., et al.2023） | 96.09 | 96.44 | 96.26 |
| DAE-NER（Sun et al., 2024) | 96.92 | 95.18 | 96.04 |
| ours | **97.16** | **96.56** | **96.86** |

OntoNotes4 NER dataset: Table 4 shows the results obtained on the OntoNotes4 dataset. The F1 score of MECT (Wu et al., 2021) which introduces character structure information is 76.92%. SoftLexicon (Ma et al., 2019) introduced incorporating lexical information into characters to enhance semantics, with an F1 score of 82.81%. CAN-NER (Zhu et al., 2019) and Token-Relation(huang et al,2022) obtained f1 scores of 73.64% and 83.28% by using an improved attention mechanism. VisPhone, which adds speech features on the basis of glyphs, achieves an F1 score of 82.63%. Compared with W²NER, the F1 value are increased by 0.17%. Our model improves recall by 1.35% compared to Token-Relation and achieves a similar F1-score.

Table 4 Results obtained on OntoNotes4.

| Models | Chinese OntoNotes4 NER | | |
|---|---|---|---|
| | Precision | Recall | F1 |
| CAN-NER(Zhu et al.,2019) | 75.05 | 72.29 | 73.64 |
| SoftLexicon (Ma et al. 2020) | 83.41 | 82.21 | 82.81 |
| MECT (Wu et al., 2021) | 77.57 | 76.27 | 76.92 |
| W²NER(Fei et al,2022) | 82.31 | 83.36 | 83.08 |
| Token-Relation(huang et al,2022) | **82.57** | 83.99 | **83.28** |
| NER-MC(Yan et al,2023) | 76.22 | 75.81 | 76.01 |
| VisPhone (Zhang, B., et al.2023） | 80.57 | 84.79 | 82.63 |
| ours | 81.25 | **85.34** | 83.25 |

MSRA NER dataset: Table 5 shows the results obtained on the MSRA dataset. Considering the importance of semantic information, the F1 scores of MECT (Wu et al., 2021) with character structure information and NER-MC(Yan et al,2023) with word boundary information are 94.32% and 93.46%, respectively. SoftLexicon (Ma et al., 2019) introduced the incorporation of lexical information into characters to enhance semantics, with an F1 score of 95.42%. CAN-NER (Zhu et al., 2019) and Token-Relation(huang et al,2022) obtained f1 scores of 92.97% and 96.13% by using an improved attention mechanism. The F1 value of VisPhone with speech features added on the basis of glyphs is 96.09%. Although our precision drops by 0.22% compared to VisPhone, our model outperforms other models in terms of both F1 score and recall.

Table 5 Results obtained on MSRA.

| Models | Chinese MSRA NER | | |
|---|---|---|---|
| | Precision | Recall | F1 |
| CAN-NER(Zhu et al.,2019) | 93.53 | 92.42 | 92.97 |
| SoftLexicon (Ma et al. 2020) | 95.75 | 95.10 | 95.42 |
| MECT (Wu et al., 2021) | 94.55 | 94.09 | 94.32 |
| W²NER(Fei et al,2022) | 96.12 | 96.12 | 96.10 |
| Token-Relation(huang et al,2022) | 96.08 | 96.18 | 96.13 |
| NER-MC(Yan et al,2023) | 94.27 | 92.66 | 93.46 |
| VisPhone (Zhang, B., et al.2023） | **96.31** | 95.83 | 96.07 |
| ours | 96.09 | **96.34** | **96.21** |

## 5.4 Ablation study

To explore the contribution of each component in the model, we evaluated the performance of the remaining components by removing each key component as follows :(1) Remove the improved Transformer coding module; (2) Remove direction and distance representations for semantic enhancement representations (3) remove all convolution (4) remove MLP or Biaffine predictors for co-prediction modules (5) remove character pair relationship enhancement modules.

Table 6 Results of model ablation experiments

| Settings | Weibo | Resume | OntoNotes4 | MSRA |
|---|---|---|---|---|
| ours | 76.63 | 96.86 | 83.25 | 96.21 |
| w/o Transformer | 70.20 | 95.30 | 79.39 | 93.99 |
| w/o Region matrix | 68.42 | 95.91 | 78.40 | 95.97 |
| w/o Distance matrix | 67.94 | 95.31 | 79.39 | 95.56 |
| w/o Dilated convolution | 74.75 | 96.07 | 82.69 | 95.99 |
| w/o MLP | 75.79 | 95.92 | 79.36 | 94.48 |
| w/o Biaffine | 75.91 | 95.80 | 79.64 | 94.58 |
| w/o Enhancement | 76.00 | 96.15 | 79.69 | 94.41 |
| w/o relation | 72.82 | 94.81 | 82.08 | 94.31 |
| (NNW, PNW, THW, HTW) | | | | |

The results of the four datasets consistently show that encoder layer BERT has the most significant impact on the model performance. Removing direction-aware and distance-aware transformers results in a significant drop, with distance information having the largest impact of 3.32% and at least 0.75% impact on four datasets. The maximum decrease of direction information is 2.61%, and it also has at least 0.54% effect. This shows that the encoder layer can better extract the context features of the text by introducing the distance and direction information between characters when enhancing semantic information. After removing all convolutions, the performance also decreases to varying degrees on different datasets, up to 1.88%, which verifies the effectiveness of multi-granularity extended convolution in capturing the relationship between characters with different distances. All experimental results show that the prediction module has the smallest performance degradation compared with other modules. The experiments are mainly discussed by removing the biaffine predictor and the MLP predictor respectively. MLP has a greater impact on the model performance than Biaffine, but the Biaffine predictor also brings at least 0.74% performance improvement on the four datasets. Finally, when the tag relationship between character pairs in the model is removed, the performance on the four data sets is significantly decreased, which indicates that the four tags relationships are effective, that is, the relationship between characters and tags is beneficial to the model's prediction of entities.

# 6. Conclusion

In this paper, we propose a Chinese NER model named CRENER, which derives a directional relative positional encoding with an unscaled self-attention mechanism adapted transformer encoder to model character-level features. Subsequently, we incorporate four predefined tags to improve the model's capacity for learning contextual semantics in Chinese NER and capturing character pair relations based on the 2D representation. Meanwhile, we use a co-predictor module to predict the final relationship among character pairs and tags. Experimental results on four well-known Chinese NER benchmark datasets show that our model performs significantly better than the baseline models of other semantic enhancement and grid tagging methods, which verifies that using the improved character encoder and relation enhancement module can effectively improve the performance of the Chinese NER. In future work, our model can be extended to more complex information extraction tasks.